%% file: coling2020.tex
\documentclass[11pt]{article}
\usepackage{coling2020}
\usepackage{times}
\usepackage{url}
\usepackage{latexsym}

\usepackage{soul, hyperref, xcolor, enumitem}
\usepackage{algorithm, algorithmicx, mathtools}
\usepackage{amsmath,amsthm,mathtools}
\usepackage{latexsym,graphicx,amssymb,multirow,booktabs}
\usepackage{makecell}
\usepackage{amsfonts}

\usepackage{subfigure}
\usepackage[font={footnotesize}]{caption}
\usepackage{microtype}
\usepackage{graphicx} 
\colingfinalcopy 

\newcommand{\mixup}{\textit{mixup}}

\title{Mixup-Transformer: Dynamic Data Augmentation for NLP Tasks}

\author{Lichao Sun$^{1}$\thanks{~ Indicates Equal Contribution}, Congying Xia{$^2$\footnotemark[1]}, Wenpeng Yin$^{3}$, Tingting Liang$^{4}$, Philip S. Yu$^{2}$, Lifang He$^{1}$ \\
  $^1$ Lehigh University;
  $^2$ University of Illinois at Chicago\\
  $^3$ Salesforce Research;
  $^4$ Hangzhou Dianzi University \\
  {\tt james.lichao.sun@gmail.com;}
  {\tt \{cxia8, psyu\}@uic.edu} \\
  {\tt wyin@salesforce.com;}
  {\tt liangtt@hdu.edu.cn} \\
  {\tt lih319@lehigh.edu}
  }

\date{}

\begin{document}
\maketitle
\begin{abstract}
  
  Mixup \cite{zhang2017mixup} is a latest data augmentation technique that linearly interpolates input examples and the corresponding labels. It has shown strong effectiveness in image classification by interpolating images at the pixel level. Inspired by this line of research, in this paper, we explore: i) how to apply \mixup\ to natural language processing tasks since text data can hardly be mixed in the raw format; ii) if \mixup\ is still effective in transformer-based learning models, \emph{e.g.}, BERT. To achieve the goal, we incorporate \mixup\ to transformer-based pre-trained architecture, named ``\mixup-transformer'', for a wide range of NLP tasks while keeping the whole end-to-end training system. 
  We evaluate the proposed framework by running extensive experiments on the GLUE benchmark. Furthermore, we also examine the performance of \mixup-transformer in low-resource scenarios by reducing the training data with a certain ratio. Our studies show that \mixup\ is a domain-independent data augmentation technique to pre-trained language models, resulting in significant performance improvement for transformer-based models.
  
\end{abstract}

\section{Introduction}
\label{intro}
Deep learning has shown outstanding performance in the field of natural language processing (NLP). Recently, transformer-based methods \cite{devlin2018bert,yang2019xlnet} have achieved state-of-the-art performance across a wide variety of NLP tasks\footnote{Please see the General Language Understanding Evaluation (GLUE) (https://gluebenchmark.com/) for tracking performance.}.
However, these models highly rely on the availability of large amounts of annotated data, which is expensive and labor-intensive.
To solve the data scarcity problem, data augmentation is commonly used in NLP tasks. For example, \newcite{wei2019eda} investigated language transformations like insertion, deletion and swap. \newcite{malandrakis2019controlled} and \newcite{yoo2019data} utilized variational autoencoders (VAEs) \cite{kingma2013auto} to generate more raw inputs. Nevertheless, these methods often rely on some extra knowledge to guarantee the quality of new inputs, and they have to be working in a pipeline. 


\newcite{zhang2017mixup} proposed \mixup, a domain-independent data augmentation technique that linearly interpolates image inputs on the pixel-based feature space. \newcite{guo2019augmenting} tried \mixup\ in  CNN \cite{lecun1998gradient} and  LSTM \cite{hochreiter1997long} for text applications. 
Despite effectiveness, they conducted \mixup\ only on the fixed word embedding level like \newcite{zhang2017mixup} did in image classification. Two questions arise, therefore: (i) how to apply \mixup\ to NLP tasks if text data cannot be mixed in the raw format? Apart from the embedding feature space, what other representation spaces can be constituted and used? ii) whether or not \mixup\ can boost the state-of-the-art further in transformer-based learning models, such as BERT \cite{devlin2018bert}.


To answer these questions, we stack a \mixup\ layer over the final hidden layer of the pre-trained transformer-based model. The resulting system can be applied to a broad of NLP tasks; in particular, it is still end-to-end trainable. 
We evaluate our proposed \mixup-transformer on the GLUE benchmark, which shows that \mixup\ can consistently improve the performance of each task.
Our contributions are summarized as follows:

\textbullet\enspace We propose the \mixup-transformer that applies \mixup\ into transformer-based pre-trained models. To our best knowledge, this is the first work that explores the effectiveness of \mixup\ in Transformer. 

\textbullet\enspace In experiments, we demonstrate that \mixup-transformer can consistently promote the performance across a wide range of NLP benchmarks, and it is particularly helpful in low-resource scenarios where we reduce the training data from 10\% to 90\%.






\section{Mixup-Transformer}
In this section, we first introduce the \mixup\ used in previous works.
Then, we show how to incorporate the \mixup\ into transformer-based methods and how to do the fine-turning on different text classification tasks.
Last, we will discuss the difference between the previous works and our new approach.

\begin{figure}[ht!]
    \centering
   \includegraphics[width=0.7\linewidth]{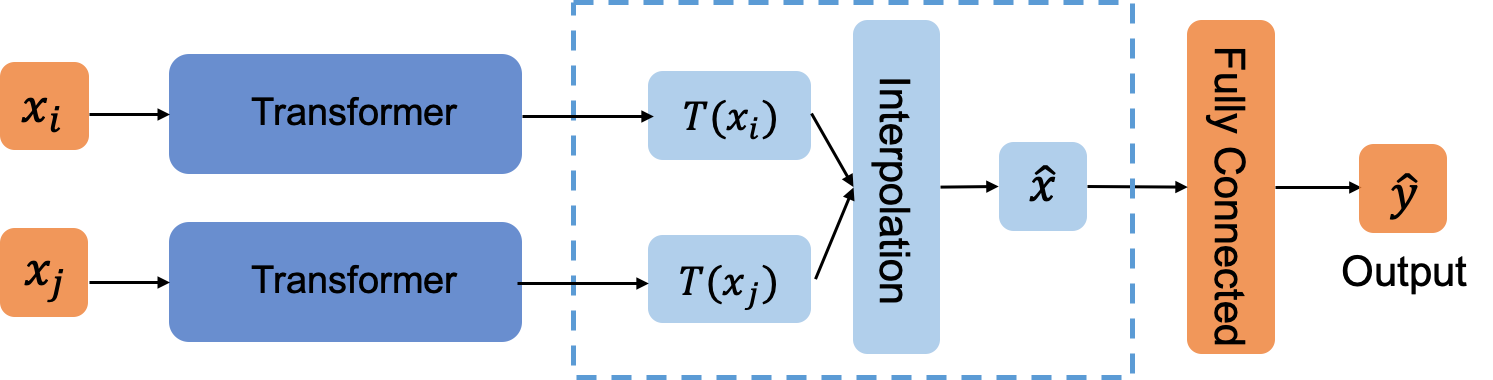}
   \caption{The overall framework of \mixup-transformer. $x_i$, $x_j$ are two separate sentences, fed to the same Transformer $T$. $T(x_i)$ is the representation of the input $x_i$, generated by $T$. $\hat{x}$ and $\hat{y}$ are the interpolated representation and label, respectively.}
     \vspace{-0.1in}
    \label{fig:framework}
\end{figure}

\subsection{Mixup}
Mixup is first proposed for image classification \cite{zhang2017mixup},
which incorporates the prior knowledge that linear interpolations of feature representations should lead to the same interpolations of the associated targets.
In {\mixup}, virtual training examples are constructed by two examples drawn at random from the training data:
\begin{align*} 
\vspace{-0.1in}
\hat{x} &= \lambda x_i + (1 - \lambda)x_j, \text{where $x_i$, $x_j$ are raw inputs;} \\ 
\hat{y} &= \lambda y_i + (1 - \lambda)y_j, \text{where $y_i$, $y_j$ are corresponding labels,}
\vspace{-0.1in}
\end{align*}
where $\lambda$ could be either fixed value in $[0, 1]$ or $\lambda \sim \text{Beta}(\alpha, \alpha)$, for $\alpha \in (0, \infty)$.
In previous works, \mixup\ is a static data augmentation approach that improves and robusts the performance in image classification.

\subsection{Mixup for Text Classification}
Text classification is the most fundamental problem in the NLP field. Unlike image data, text input consists of discrete units (words) without an inherent ordering or algebraic operations - it could be one sentence, two sentences, a paragraph or a whole document.

The first step of text classification is to use the word embedding to convert each word of the text into a vector representation.
In the traditional approaches, the word embedding method can be bag-of-words, or a fixed word to vector mapping dictionary built by CNN or LSTM.
In our approach, instead of using the traditional encoding methods,
we use transformer-based pre-trained language models to learn the representations for text data.
For downstream tasks, we fine-tune transformer-based models with the \mixup\ data augmentation method.
Formally, \mixup-transformer constructs virtual hidden representations dynamically durning the training process as follows:
\begin{align*} 
\vspace{-0.05in}
\hat{x} &= \lambda \cdot T(x_i) + (1 - \lambda) \cdot T(x_j), \text{where $T(x_i)$, $T(x_j)$ are output features from transformers;} \\ 
\hat{y} &= \lambda y_i + (1 - \lambda)y_j, \text{where $y_i$, $y_j$ are corresponding labels,}
\vspace{-0.1in}
\end{align*}
where $T(\cdot)$ represents outputs of the transformer layers as shown in Figure \ref{fig:framework}. Note that, the mixup process is trained together with the fine-tuning process in an end-to-end fashion, and the hidden \mixup\ representations are dynamic during the training process.

\subsection{Discussion}
In this section, we highlight two main differences between our approach and previous methods using \mixup\ techniques \cite{zhang2017mixup,guo2019augmenting} for comparison.

\textbullet\enspace {Dynamic \mixup\ representation}.
For each input pair, $x_i$ and $x_j$, the previous approaches produce a fixed \mixup\ representation given a fixed $\lambda$.
However, the \mixup\ hidden representations in our approach are dynamic since they are trained together with the fine-tuning process.

\textbullet\enspace {Dynamic \mixup\ activation}.
Since a pre-trained network needs to be fine-tuned for a specific task, we can dynamically activate the \mixup\ during the training.
For example, if the training epoch is 3, we can choose to use \mixup\ in any epoch or all epochs.
In our experiments, we fine-tune the model without \mixup\ in the first half of epochs for good representations and add \mixup\ in the last half of the epochs.

\input{exp.tex}

\section{Conclusion and Future Work}
In this paper, we propose the \mixup-transformer that incorporates a data augmentation technique called \mixup\ into transformer-based models for NLP tasks.
Unlike using the static \mixup\ in previous works, our approach can dynamically construct new inputs for text classification.
Extensive experimental results show that \mixup-transformer can be dynamically used with a pre-trained model to achieve better performance on GLUE benchmark.
Two future directions are worth considering on text data. First, how to use \mixup\ on other challenging NLP problems, such as zero-shot, few-shot or meta-learning tasks. Second, how to do \mixup\ on document-level text data like paragraphs.
Instead of using \mixup\ directly, we may need to extract appropriate information from the data in the training process.
Selecting the right information to mix up for text classification would be an exciting and challenging area to be in.

\bibliographystyle{coling}
\bibliography{coling2020}

\end{document}

%% file: exp.tex
\section{Experiments}
To show the effectiveness of our proposed \mixup-transformer, we conduct extensive experiments by adding the \mixup\ strategy to transformer-based models on seven NLP tasks contained in the GLUE benchmark. Furthermore, we reduce the training data with different ratios (from 10\% to 90\%) to see how the \mixup\ strategy works with insufficient training data. We report the performance on development sets for all the tasks because the test time is limited by the online GLUE benchmark.

\paragraph{Datasets.}
The General Language Understanding Evaluation (GLUE) benchmark \cite{wang2018glue} is a collection of diverse natural language understanding tasks. Experiments are conducted on eight tasks in GLUE: CoLA \cite{warstadt2019neural}, SST-2 \cite{socher2013recursive}, MRPC \cite{dolan2005automatically}, STS-B \cite{cer2017semeval}, QQP \cite{qqp}, MNLI \cite{williams2017broad}, QNLI \cite{rajpurkar2016squad}, RTE \cite{bentivogli2009fifth}.

\paragraph{Baselines.}
Two baselines are conducted in the experiments, including BERT-base and BERT-large \cite{devlin2018bert}. We evaluate the performance of our methods by adding the \mixup\ strategy to these two baselines.

\paragraph{Implementation details.}
When fine-tuning BERT with or without the \mixup\ strategy for these NLP tasks, we fix the hyper-parameters as follows: the batch size is 8, the learning rate is 2e-5, the max sequence length is 128, and the number of the training epochs is 3.
We test different values of $\lambda$, (from 0.1 to 0.9) on the default dataset (CoLA) and find \mixup-transformer is insensitive to this hyper-parameter, so we set a fixed value of $\lambda = 0.5$.

\subsection{Results on full data}

\begin{table*}[ht!]
\centering
\resizebox{6in}{!}{
\begin{tabular}{l|cccccccc}
\Xhline{3\arrayrulewidth}
\multirow{2}{*}{Dataset}  & CoLA & SST-2   & MRPC  & STS-B  & QQP   & MNLI-mm & QNLI    & RTE     \\ 
& (Corr)	& (Acc) & 	(Acc) &	(Corr) & (Acc)& 	(Acc)& 	(Acc)& 	(Acc) \\\Xhline{3\arrayrulewidth}
\multicolumn{1}{l|}{BERT-base} &          57.86                        &       92.20                       &         86.76                   &      \textbf{89.41}                       &                     90.79        &                 84.47          &         91.61                &      68.23      \\ 
\multicolumn{1}{l|}{BERT-base + \mixup\ } &    \textbf{59.58}                             &      \textbf{92.78}                        &       \textbf{88.48}         &   88.66                   &       \textbf{90.98}                   &                  \textbf{85.12}          &                   \textbf{91.84}     &         \textbf{71.84}           \\ 
Improved & 1.72 & 	0.58 & 	1.72 & -0.75 & 	0.21 & 0.65 & 	0.23& 	3.61\\
\hline
\multicolumn{1}{l|}{BERT-large} &          59.71                        &       91.97                       &            86.27                   &      89.21                        &                      90.38        &                  \textbf{85.92}           &           91.96                     &      69.31       \\ 
\multicolumn{1}{l|}{BERT-large + \mixup\ } &     \textbf{62.39}                             &      \textbf{92.78}                        &        \textbf{87.99}                      &     \textbf{90.10}                         &       \textbf{90.80}                      &                  85.91           &                        \textbf{92.09}      &                \textbf{69.68}           \\ 
Improved & 2.68 & 	0.81 & 	1.72 & 0.89 & 	0.42 & 	-0.01 &	0.13 & 	0.37 
\\ \Xhline{3\arrayrulewidth}
\end{tabular}
}
\caption{\mixup-transformer results on eight NLP tasks. Matthew's Correlations are reported for CoLA, Spearman correlations are reported for STS-B, and
accuracy scores are reported for the other tasks.}
\label{exp1}
\end{table*}

Experimental results for eight different NLP tasks are illustrated in Table \ref{exp1}. By adding the proposed {\mixup} technique to BERT-base and BERT-large, the \mixup-transformer improves the performance consistently on most of these tasks. The average improvement is around 1\% for all the settings. The highest performance gain comes from the RTE task by adding the proposed {\mixup} technique on BERT-base. In this experiment, the accuracy improves from 68.23\% to 71.84\%, which is an increase of 3.61\%. The Matthew's correlation for CoLA increases from 59.71\% to 62.39\% (improved 2.68\%) with {\mixup} on BERT-large. 
Some experiments also get performance decrease with {\mixup}. For example, adding {\mixup} to BERT-base on STS-B decreases the Spearman correlation from 89.41\% to 88.66\%. 
Overall, most of the tasks improved (14 out of 16, while 2 got slightly worse) were found by applying \mixup-transformer.

\begin{table*}[ht!]
\centering
\resizebox{\linewidth}{!}{
\begin{tabular}{l|cccccccccc}
\Xhline{3\arrayrulewidth}
Training Data    & 10\%    & 20\%    & 30\%    & 40\%    & 50\%    & 60\%    & 70\%    & 80\% & 90\%    & 100\% \\ \hline
BERT-large       & 74.51 & 77.45 & 77.69 & 82.11 & 83.58 & 84.07 & 84.07 & 84.56 & 86.76 & 85.53  \\
BERT-large + \mixup\ & \textbf{77.21} & \textbf{79.17} & \textbf{81.13} & \textbf{87.01} & \textbf{85.78}& \textbf{86.27} & \textbf{86.79} & \textbf{86.76} & \textbf{87.25} & \textbf{87.99}  \\ \hline
Improved          & 2.70  & 1.72  & 3.44  & 4.90 & 2.21 & 2.21  & 2.72 & 2.21  & 0.49  & 2.46 \\
\Xhline{3\arrayrulewidth}
\end{tabular}
}
\caption{\mixup-transformer with reduced training data on MRPC. Accuracy scores are used to evaluate the performance.}
\label{exp2}
\end{table*}

\subsection{Results on limited data}

\begin{figure}[ht]
\centering
\subfigcapskip=-5pt
\subfigure[STS-B]{\includegraphics[width=0.45\linewidth]{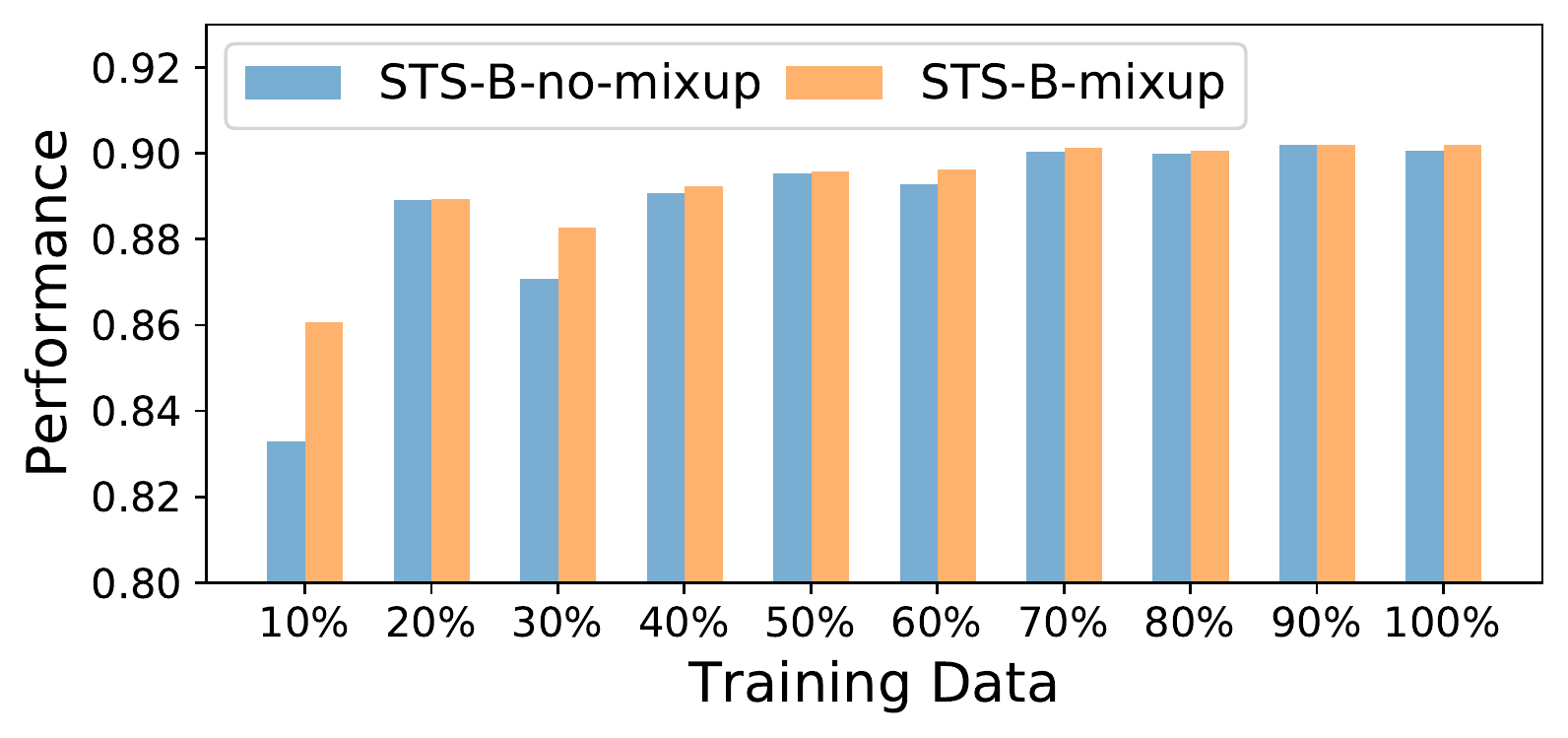}}
\vspace{-0.1in}
\subfigure[MRPC]{\includegraphics[width=0.45\linewidth]{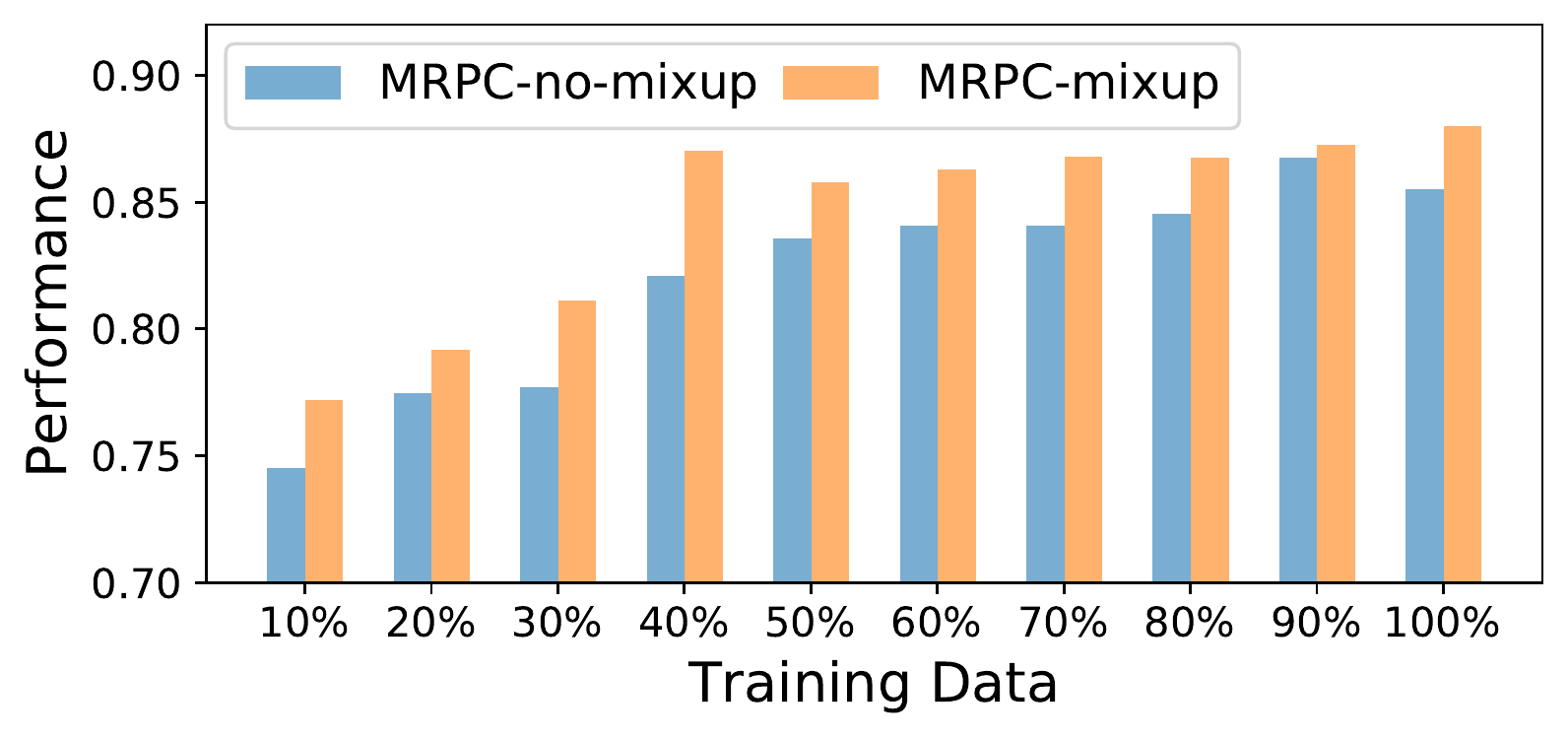}}
\vspace{-0.1in}
\subfigure[RTE]{\includegraphics[width=0.45\linewidth]{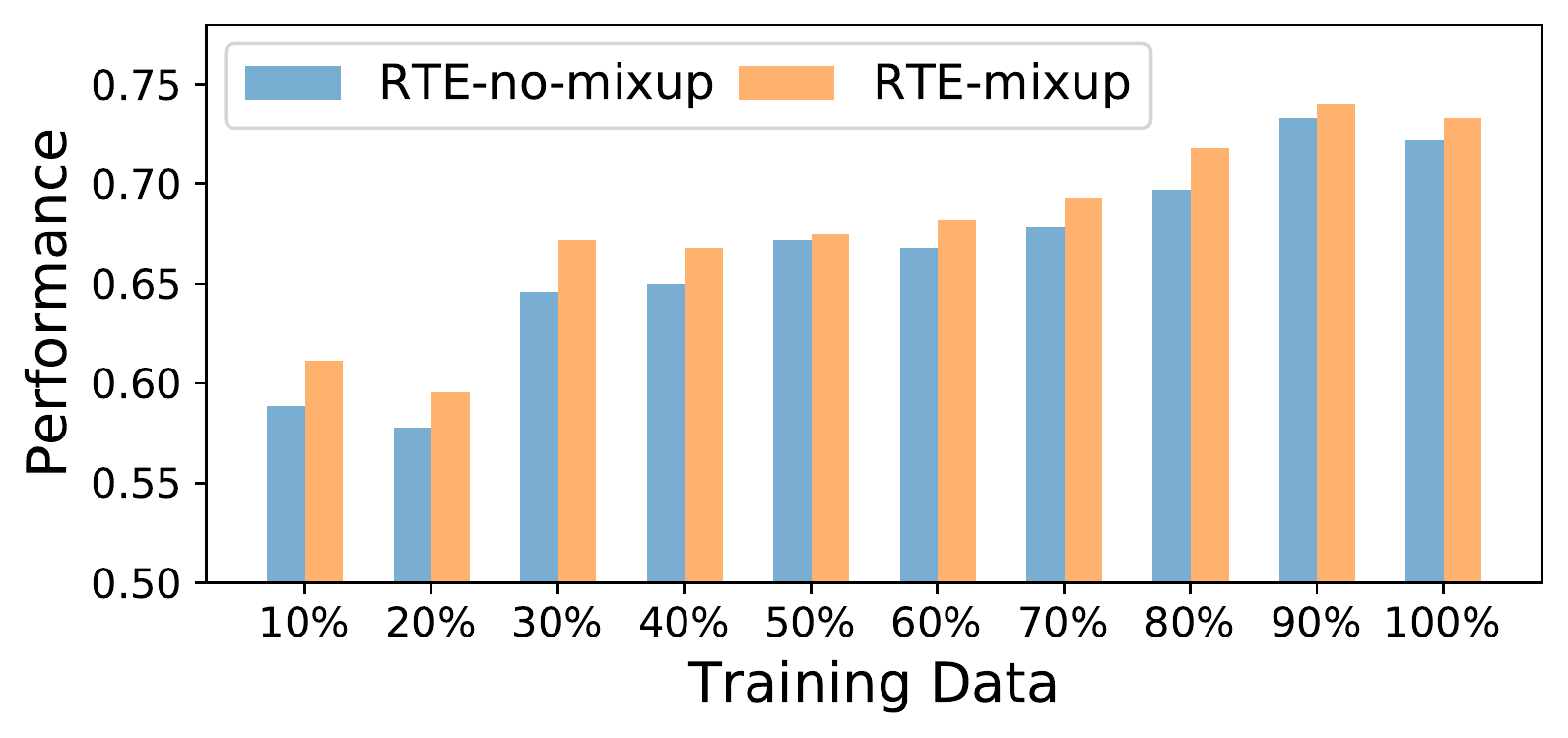}}
\subfigure[CoLA]{\includegraphics[width=0.45\linewidth]{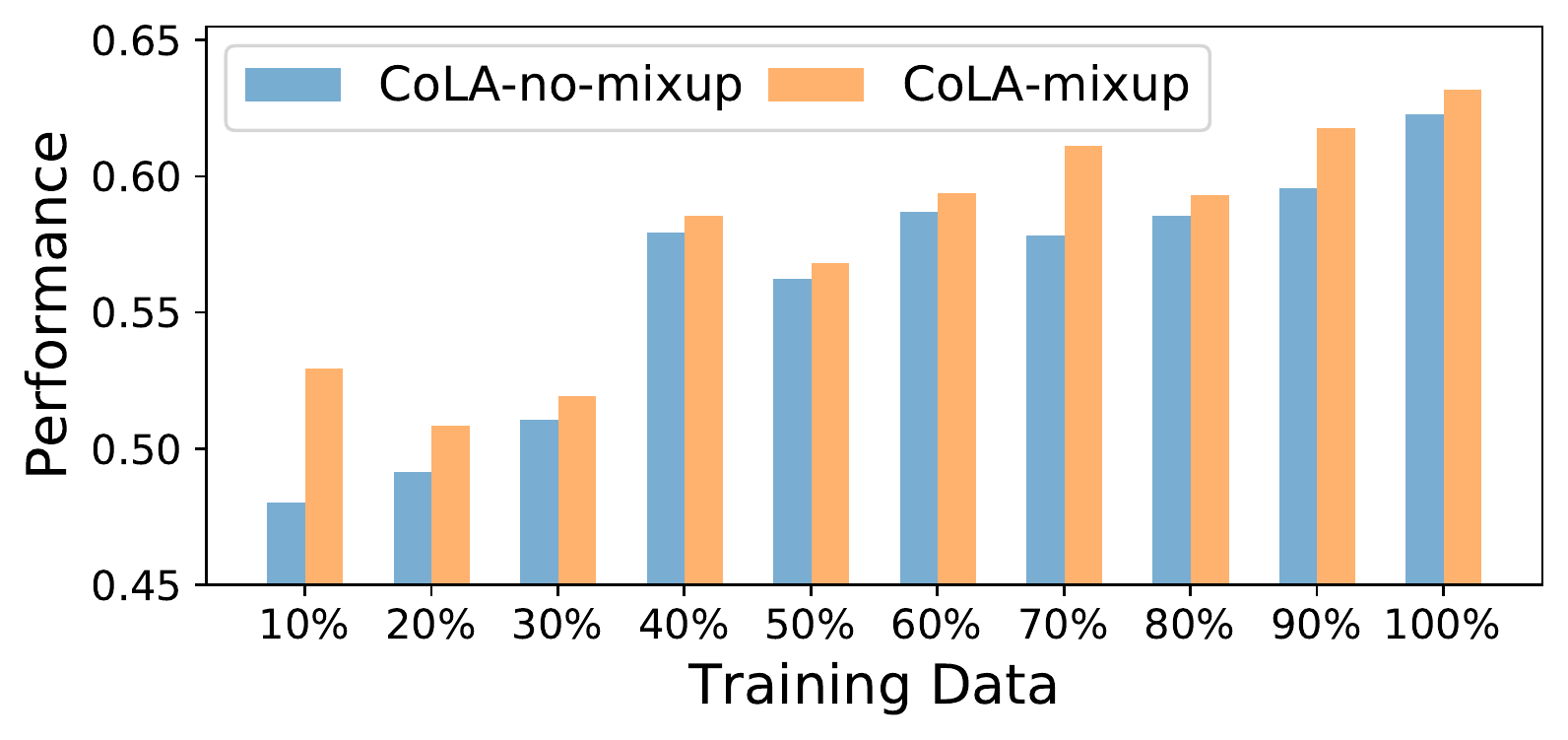}}
\caption{\mixup-transformer with BERT-large runs with reduced training data for four tasks: STS-B, MRPC, RTE and CoLA.}
\label{fig:small}
\end{figure}

As {\mixup} is a technique for augmenting the feature space, it is interesting to see how it works when the training data is insufficient. Therefore, we reduce the training data with a certain ratio (from 10\% to 90\% with a step 10\%) and test the effectiveness of {\mixup}-transformer in low-resource scenarios. As shown in Table 2, BERT-large + \mixup\ consistently outperforms BERT-large when we reduce the training data for MRPC, where the highest improvement (4.90\%) is achieved when only using 40\% of the training data. Using the full training data (100\%) gets an increase of 2.46\%, which indicates that {\mixup}-transformer works even better with reduced annotations. We also report the experiments of reducing training data on other tasks, including STS-B, RTE and CoLA. As shown in Figure \ref{fig:small}, the {\mixup}-transformer again consistently improves the performance for all the experiments. 
The performance gains with less training data (like 10\% for STS-B, CoLA, and RTE) are higher than using full training data since data augmentation is more effective when the annotations are insufficient.
Therefore, the {\mixup} strategy is highly helpful in most low-resource scenarios.

